\newcommand{\val}{\textit{valence}}
\newcommand{\aro}{\textit{arousal}}
\newcommand{\dom}{\textit{dominance}}
\newcommand{\Val}{\textit{Valence}}
\newcommand{\Aro}{\textit{Arousal}}

\newcommand{\va}{\val{} and \aro{}}
\newcommand{\VA}{\Val{} and \Aro{}}
\newcommand{\affectnet}{AffectNet}
\newcommand{\emotic}{EMOTIC}

\newcommand\copyrighttext{%
  \footnotesize \textcopyright 2024 IEEE.  Personal use of this material is permitted.  Permission from IEEE must be obtained for all other uses, in any current or future media, including reprinting/republishing this material for advertising or promotional purposes, creating new collective works, for resale or redistribution to servers or lists, or reuse of any copyrighted component of this work in other works.}
\newcommand\copyrightnotice{%
\begin{tikzpicture}[remember picture,overlay]
\node[anchor=south,yshift=10pt] at (current page.south) {\fbox{\parbox{\dimexpr\textwidth-\fboxsep-\fboxrule\relax}{\copyrighttext}}};
\end{tikzpicture}%
}
\documentclass[10pt,twocolumn,letterpaper]{article}

\usepackage[pagenumbers]{cvpr} % To force page numbers, e.g. for an arXiv version

\usepackage{tabularx,booktabs}
\usepackage[accsupp]{axessibility} % Improves PDF readability for those with disabilities.

\usepackage[dvipsnames]{xcolor}

\definecolor{cvprblue}{rgb}{0.21,0.49,0.74}
\usepackage[pagebackref,breaklinks,colorlinks,citecolor=cvprblue]{hyperref}
\usepackage{tikz}
\def\paperID{33} % *** Enter the Paper ID here
\def\confName{CVPR}
\def\confYear{2024}

\title{CAGE: Circumplex Affect Guided Expression Inference}

\author{Niklas Wagner$^{1}$$^,$$^*$, Felix Mätzler$^{1}$$^,$$^*$, Samed R. Vossberg$^{1}$$^,$$^*$, Helen Schneider$^{1}$$^*$, Svetlana Pavlitska$^{2}$, \\J. Marius Zöllner$^{1,2}$\\
\textit{$^{1}$ Karlsruhe Institute of Technology (KIT), Germany}\\
\textit{$^{2}$ FZI Research Center for Information Technology, Germany} \\
{\tt\small helen.schneider@kit.edu}\\
}
\begin{document}
\maketitle
\def\thefootnote{*}\footnotetext{These authors contributed equally to this work}
\copyrightnotice
\thispagestyle{empty}
\pagestyle{empty}
\begin{abstract}
Understanding emotions and expressions is a task of interest across multiple disciplines, especially for improving user experiences. Contrary to the common perception, it has been shown that emotions are not discrete entities but instead exist along a continuum. People understand discrete emotions differently due to a variety of factors, including cultural background, individual experiences, and cognitive biases. Therefore, most approaches to expression understanding, particularly those relying on discrete categories, are inherently biased. In this paper, we present a comparative in-depth analysis of two common datasets (\affectnet{}  and \emotic{}) equipped with the components of the circumplex model of affect. Further, we propose a model for the prediction of facial expressions tailored for lightweight applications. Using a small-scaled MaxViT-based model architecture, we evaluate the impact of discrete expression category labels % (\textit{Neutral, Happiness, Sadness, Surprise, Fear, Disgust, Anger, Contempt}) 
in training with the continuous \va{} labels. We show that considering valence and arousal in addition to discrete category labels helps to significantly improve expression inference.  The proposed model outperforms the current state-of-the-art models on \affectnet{}, establishing it as the best-performing model for inferring \va{} achieving a 7\% lower RMSE. Training scripts and trained weights to reproduce our results can be found here: \url{https://github.com/wagner-niklas/CAGE_expression_inference}.
\end{abstract}    
\section{Introduction}
\label{sec:intro} 

The inference of emotions through expressions has been a topic of interest for the past years as it might give insights into a person's feelings towards other individuals or topics. Mehrabian and Wiener~\cite{mehrabian1967decoding} suggest 55\% of communication is perceived by expressions. Lapakko~\cite{Lapakko2015CommunicationI9} argues, however, that these findings are limited to emotional states. Automation of analysis of expressions to get insights into user experience is one step towards live feedback without direct interaction with an individual. 

\begin{figure}[t]
    \centering
    \includegraphics[width=0.8\columnwidth, trim={3cm 11cm 3cm 3cm}, clip]{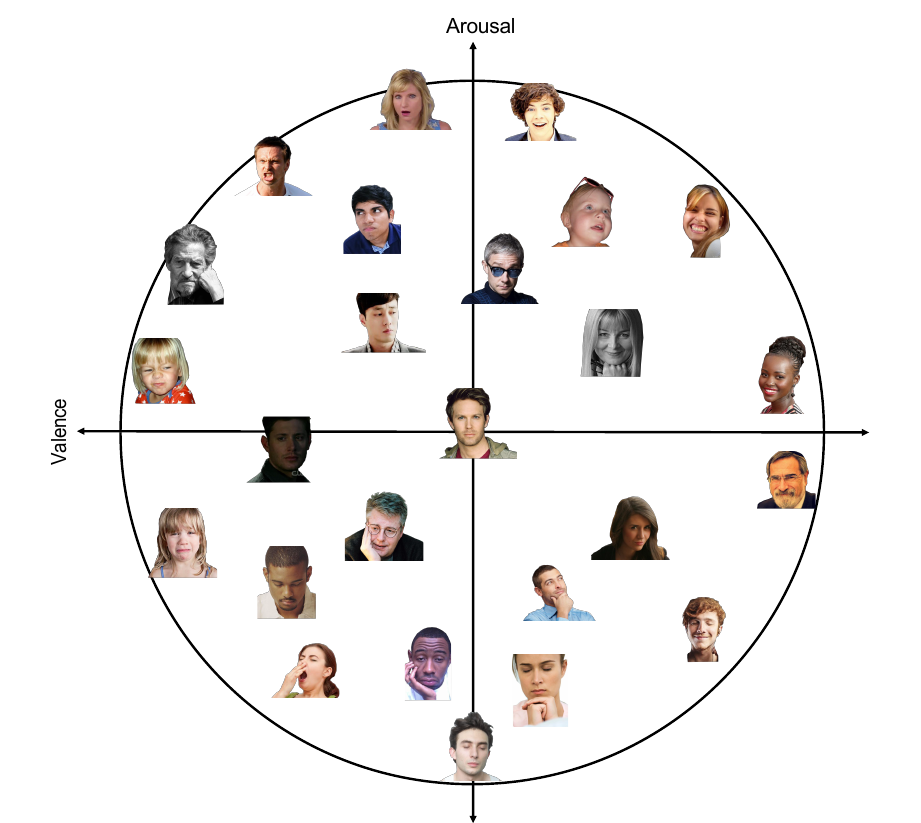}
    \caption{\textit{Valence/arousal} for sample images from \affectnet{}~\cite{mollahosseini2017affectnet}.}
    \label{fig:Russel_Affectnet}
\end{figure}

A common approach is \textit{expression inference}, i.e. classification of emotional expressions into discrete categories. However, a comprehensive meta-analysis of facial expressions research by Barrett et al.~\cite{barretetal2019}  has shown, that there is no consensus across cultures and intra-cultural over specific facial movements reliably depicting one category of emotion. They suggest that affective states can more reliably be inferred by a third-party individual. They emphasize that these states are inferred, not recognized. According to Russell~\cite{rusellmodell}, affects can be described as a set of dimensions with each dimension varying independently. These dimensions are called \va{}, representing the positivity/negativity and intensity-/activation of expressions respectively. Using \va{} of the circumplex model of affect~\cite{rusellmodell} as additional dimensions rather than only discrete emotions for expression inference thus offers a more robust framework, as they provide a continuous spectrum that captures the underlying affective states.

In this work, we compare training with \va{} labels merged with the commonly used discrete emotions to train with the two approaches separately. 
Our approach involves pinpointing the differences and similarities between two leading datasets that catalog images according to their explicit discrete and continuous emotional states: \affectnet~\cite{mollahosseini2017affectnet} and \emotic{}~\cite{kosti_emotic_2017}. 
% We examine the labeling process and show that there is still a need for further datasets to create a universal model that does not guess emotions based on labels of third-party individuals but rather gets information about the true internal state of each image subject.
% We refer to~\cite{barretetal2019} for a more detailed discussion. 
We then develop a lightweight deep neural network tailored for computer vision tasks, aiming to accurately infer these discrete emotions as well as the continuous dimensions of \va{}, surpassing the performance of existing models. In particular, our model improves accuracy by reducing the root-mean-square error  (RMSE) by 7.0\% for \val{} and 6.8\% for \aro{}. It also increases the concordance correlation coefficients (CCC) by 0.8\% for \val{} and 2.0\% for \aro{} when tested on the \affectnet{} dataset. These improvements are reflected in our final results, with CCC values of 0.716 for \val{} and 0.642 for \aro{}, and RMSE values of 0.331 for \val{} and 0.305 for \aro{}. Furthermore, we exceed the top-3 accuracy set by Khan \etal~\cite{khan2024focusclip} on the \emotic{} dataset by 1.0\%. %Additionally, we conduct a cross-evaluation of the model's effectiveness using the given test datasets. %In summary, this research focuses on the following question:
% Consequently, the impact of these approaches needs to be measured. To do so, we 
% \begin{enumerate} [label=(\roman*)]
%     \item identify differences and similarities of two state-of-the-art datasets containing images with their apparent discrete and continuous emotion states
%     \item enhance a lightweight deep neural network architecture suited for computer vision to suggest these discrete emotions and/or continuous dimensions \va{}l{}, \aro{}
%     \item cross-evaluate the resulting model performances on the given test datasets. Hence, the following research question motivates our research:
% \end{enumerate}
%\textit{What effect does the addition of \val{}/\aro{} regression to discrete emotion classification have on facial emotion guessing performance across datasets?} 

%In the following we examine related work in \autoref{sec:relatedwork}, looking into different datasets and related approaches in emotion guessing. Then, we describe our applied data analysis and our model training approach in \autoref{sec:method}. Subsequently, we share our insights on the data and the outcomes of our model training in \autoref{sec:results}. Lastly, we present our conclusions and future perspectives in \autoref{sec:conclusion}.

% \newpage
% \clearpage
\section{Related Work}
\label{sec:relatedwork}

% In this chapter, we go into the existing research and methodologies relevant to emotion guessing via facial expressions. 
In the field of affective computing, in particular expression inference, the integration of \val{}/\aro{} regression with discrete emotion classification has emerged as a promising approach to enhance the performance and applicability across diverse datasets. In the following, we discuss existing works in this domain.

\subsection{Datasets for Expression Inference}

In the domain of expression inference, several datasets exist. However, these datasets vary significantly in both the data they offer and their popularity.
Among the most widely used datasets are FER2013~\cite{goodfellow_challenges_2013} and FERPlus~\cite{barsoum_training_2016}, which provide annotated 48$\times$48 pixel black-and-white facial images classified in seven (FER) or eight (FER+) discrete emotional states. 
% While these datasets have contributed significantly to the advancement of emotion state research, they may have limitations in capturing the complexity and nuances of human emotions due to limited data labeling. In our approach, we have therefore chosen the \affectnet{}~\cite{mollahosseini2017affectnet} and \emotic{}~\cite{emotic_pami2019} datasets. 
While these datasets have been the foundation for numerous research contributions, they have been expanded in various ways over the past years. Notable examples in this context are the \emotic{}~\cite{kosti_emotic_2017} and \affectnet{}~\cite{mollahosseini2017affectnet} datasets, which both contain high-resolution RGB images.
\affectnet{} is a large-scale database containing around 0.4 million facial images labeled by 12 annotators. Each image is annotated with categorical emotions, mirroring those used in the FER+ dataset, in addition to \va{}  values. This approach offers a more refined representation of emotions compared to categorical labels only. 

The \emotic{} (\textit{Emotions in Context}) dataset provides a more nuanced perspective on affective states. Unlike earlier datasets focused solely on facial expressions, \emotic{} captures individuals in full-body shots within their surrounding context. \emotic{}  features bounding boxes that encompass each individual's entire body, eliminating the need for a visible face. Furthermore, it categorizes emotions into 26 discrete categories, allowing for multiple labels per individual. In addition, the dataset expands these discrete values with continuous measures of \va{} as well as \dom{} that measures the level of control a person feels during a situation, ranging from submissive / non-control to dominant / in-control~\cite{emotic_pami2019}. 

While there are at least 28 datasets such as CK+~\cite{5543262}, RAF-DB~\cite{Li_2017_CVPR} or Aff-Wild2~\cite{kollias2023abaw2, kollias2023multi, kollias2022abaw, kollias2023abaw, kollias2021analysing, kollias2021affect, kollias2021distribution, kollias2020analysing, kollias_expression_2019, kollias2019deep, kollias2019face, zafeiriou2017aff, kollias2019affwild2} focusing specifically on \textit{facial expression recognition/inference} featuring continuous and/or discrete measures, we chose to focus on the two mentioned above, since we are interested in both discrete emotion labeling on an individual basis as well as continuous measures of \va{}.
\affectnet{}~\cite{mollahosseini2017affectnet} as a state-of-the-art, is arguably the most represented dataset in the current research field. 
On the other hand, \emotic{}, although not being the most utilized dataset, offers the most refined representation of measures while still focusing on a combination of discrete and continuous variables to define individuals emotion.
% Außerdem hier noch Related work angeben was schon gemacht wurde im Sinne vergleich? Gibt es Paper die unser Thema schon genauer anschauen? Gibt es Vergleich zwischen den Datensätzen oder zwischen FER und den einzelnen?! Hier einbinden 
% Elicit research machen

\begin{table}[t]
\centering
\begin{tabular}{r | c  | c }
\hline
\textbf{Method} & \textbf{Accuracy [\%]} & \textbf{Date [mm-yy]} \\
\hline
DDAMFN~\cite{electronics12173595} & 64.25 & 08-23  \\
POSTER++~\cite{mao2023poster} &63.77 & 01-23   \\
S2D~\cite{chen2023static}&63.06 & 12-22   \\
MT EffNet-B2~\cite{9815154} & 63.03 & 07-22   \\
MT-ArcRes~\cite{kollias_expression_2019} & 63.00 & 09-19   \\ \hline
\end{tabular}
\caption{Top five models on \affectnet{}-8 benchmark~\cite{paperswithcodeaff}.}
\label{tab:relatedworkaffectnet8}
\end{table}

\begin{table}[t]
\centering
\begin{tabular}{r | c  | c }
\hline
\textbf{Method} & \textbf{Accuracy [\%]} & \textbf{Date [mm-yy]} \\
\hline
S2D~\cite{chen2023static}&67.62 & 12-22   \\
POSTER++~\cite{mao2023poster} &67.49 & 01-23   \\
DDAMFN~\cite{electronics12173595} & 67.03 & 08-23  \\
Emo\affectnet{}~\cite{RYUMINA2022435} & 66.49 & 12-22  \\
Emotion-GCN~\cite{Antoniadis_2021} & 66.46 & 07-21 \\\hline
\end{tabular}
\caption{Top five models on \affectnet{}-7 benchmark~\cite{paperswithcodeaff}.}
\label{tab:relatedworkaffectnet7}
\end{table}

\subsection{Expression Inference Models}

Expression inference on datasets like \affectnet{} has been addressed in numerous publications. 
According to Paperswithcode~\cite{paperswithcodeaff}, 207 \affectnet{}-related papers have been published since 2020. Tables~\ref{tab:relatedworkaffectnet8} and~\ref{tab:relatedworkaffectnet7} show five best models in leaderboards for the \affectnet{}-8 and \affectnet{}-7 test benchmark as of 01.01.2024. As the initial FER dataset does not contain the emotion \textit{Contempt}, there exists also an \affectnet{}-7 benchmark omitting this emotion. 
So far, the best-performing models for expression inference have been almost exclusively based on convolutional neural networks (CNNs), e.g. ResNet-18~\cite{he2016deep}. Although CNNs are still competitive as shown by Savchenko \etal~\cite{9815154}, more recent architectures like the POSTER++~\cite{mao2023poster} facilitate hybrid facial expression inference via networks that combine CNNs for feature extraction with vision transformer elements for efficient multi-scale feature integration and attention-based cross-fusion, achieving state-of-the-art performance with reduced computational cost.
Because \emotic{} allows for multiple discrete labels for each individual, a general accuracy score is less applicable. Instead, Khan \etal~\cite{khan2024focusclip} suggests the \textit{top-k accuracy} can provide more insights. Utilizing a multi-modal approach leveraging region of interest heatmaps, a vision encoder, and a text encoder they achieve a top-3 accuracy of 13.73\%. 
% Far less popular is \emotic{}, cited by 63 papers since 2020 according to Paperswithcode (fig.~\ref{tab:relatedworkemotic})
% \begin{table}[htbp]
% \centering
% \begin{tabular}{r | c  | c }
% \textbf{Method} & \textbf{mAP} & \textbf{Date [mm-yy]} \\
% \hline
% EmotiCon~\cite{mittal2020emoticon} &35.48 &  03-20 \\
% EmotiCon (GCN)~\cite{mittal2020emoticon} & 32.03 &  03-20 \\
% Fusion Model 1~\cite{Kosti_2019} & 29.45 &  03-20 \\
% Fusion Model 2~\cite{Kosti_2019} & 27.70 & 03-20 \\
% CAER-Net~\cite{Lee_2019_ICCV} & 20.84 & 10-19 \\
% \end{tabular}
% \caption{Comparison Top-5 \emotic{} Benchmarks~\cite{paperswithcodeemo} }
% \label{tab:relatedworkemotic}
% \end{table}
Khor Wen Hwooi \etal~\cite{hwooi_deep_2022} suggested to extract features from CNNs and then apply model regression with a CultureNet~\cite{rudovic2018culturenet} for the continuous prediction of affect from facial expression images within the \va{} space. The best results were achieved with DenseNet201~\cite{huang2017densely} for feature extraction. The work demonstrates superior performance in predicting \va{} levels, particularly on the \affectnet{} dataset.
%The authors highlight their model's ability to generalize across unseen datasets by testing on the Aff-Wild2~\cite{kollias_expression_2019} dataset.

% \newpage
% \clearpage
\section{Analysis of Datasets for Inference of Emotional Expressions}
\label{sec:datasets}

%In this chapter, we outline the methodology employed for the comparative analysis of the \affectnet{} and \emotic{} datasets.

% \subsection{Comparing Datasets}

% To assess the two datasets, we conducted a comparative analysis to identify similarities and dissimilarities in terms of:
% \begin{itemize}
% \item Annotation granularity and emotion categories
% \item Data distribution and imbalance
% \item Annotation reliability and consistency within and between the datasets
% \end{itemize}

% Through our comparative analysis, we aimed to identify key differences between the \affectnet{} and \emotic{} datasets, which may impact the performance and generalization capability of emotion recognition models trained on these datasets.

% ********************************
% ** Subsection about the dataset comparison **
% ********************************

\begin{figure}[t]
\centering
\hspace{-1cm}
\begin{subfigure}{0.6\linewidth}
  \includegraphics[width=\linewidth]{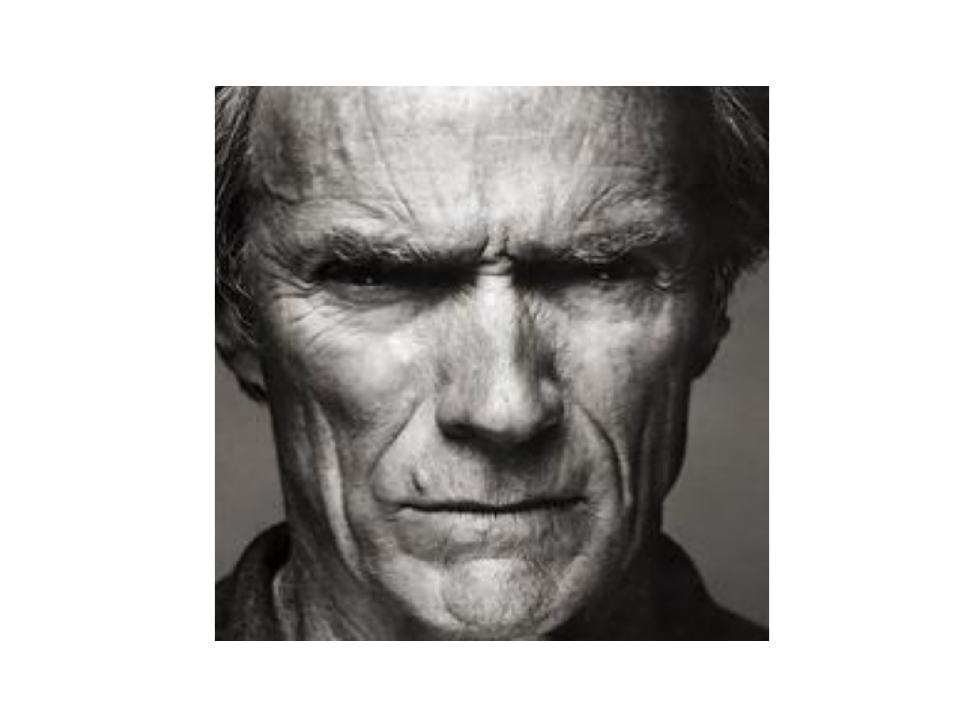} 
    \caption{AffectNet-8 (\textit{anger})}
    \label{body}
    \end{subfigure} \hspace{-1cm}
\begin{subfigure}{0.6\linewidth}
    \includegraphics[width=\linewidth]{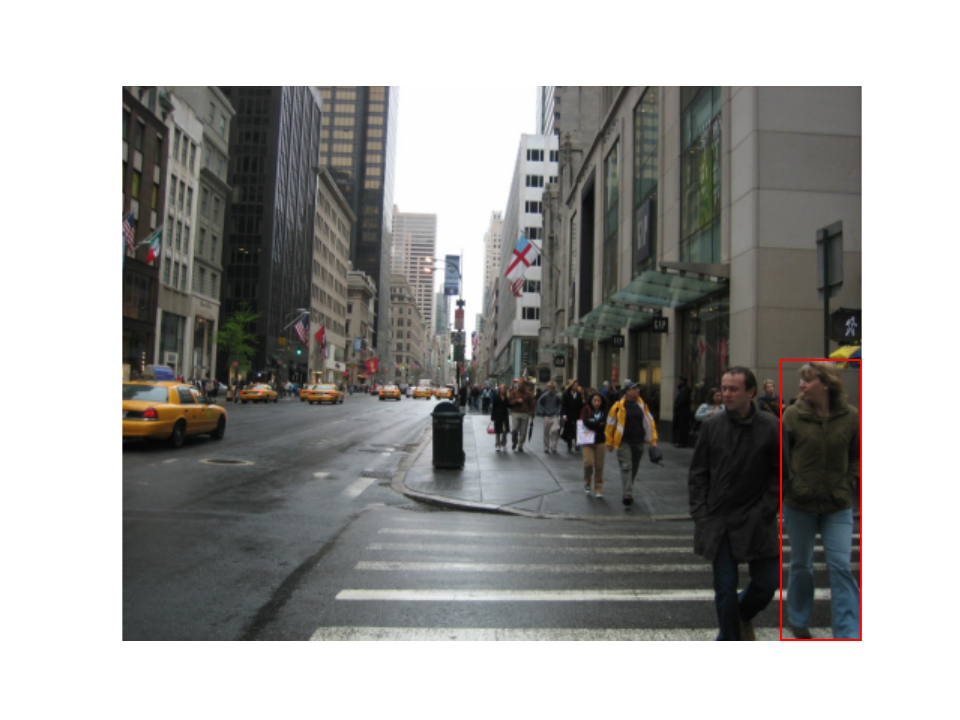}
        \caption{EMOTIC (\textit{disconnection})}
    \end{subfigure}
    \caption{Example images from AffectNet-8 and EMOTIC.}
    \label{fig:example-imgs}
\end{figure}

\begin{table}[t]
\centering
\begin{tabular}{ r | c | c }
\hline
 \textbf{Property} & \textbf{\affectnet{}-8} & \textbf{\emotic{}} \\ 
\hline
 Train Images & 287,651 & 23,266 \\
 Validation Images & 0 & 3,315 \\
 Test Images & 3,999 & 7,203 \\
 Distinct Expressions & 8 & 26 \\
 Valence & $\checkmark$ & $\checkmark$ \\
 Arousal & $\checkmark$ & $\checkmark$ \\
 Dominance & $\times$ & $\checkmark$\\
 Scale for valence, & [-1, 1] (floats) & [1, 10] (integers) \\
 arousal, dominance & & \\ \hline
\end{tabular}
\caption{Comparison of \affectnet{}-8 and \emotic{} datasets.}
\label{tab:dataset_properties}
\end{table}

%\subsection{Comparison of Datasets}
We assessed the predictive capabilities of \affectnet{} and \emotic{} (see Table~\ref{tab:dataset_properties}), rating the dataset size, expression quantity, and the encoded dimension of the circumplex model. \emotic{} dataset has a much smaller data size whilst containing several more discrete expressions and offering the additional continuous value \dom{} in comparison to the \affectnet{}-8 dataset.  In the following, we provide an in-depth analysis of the two datasets.

\begin{figure}[t]
    \centering
    \includegraphics[width=\columnwidth]{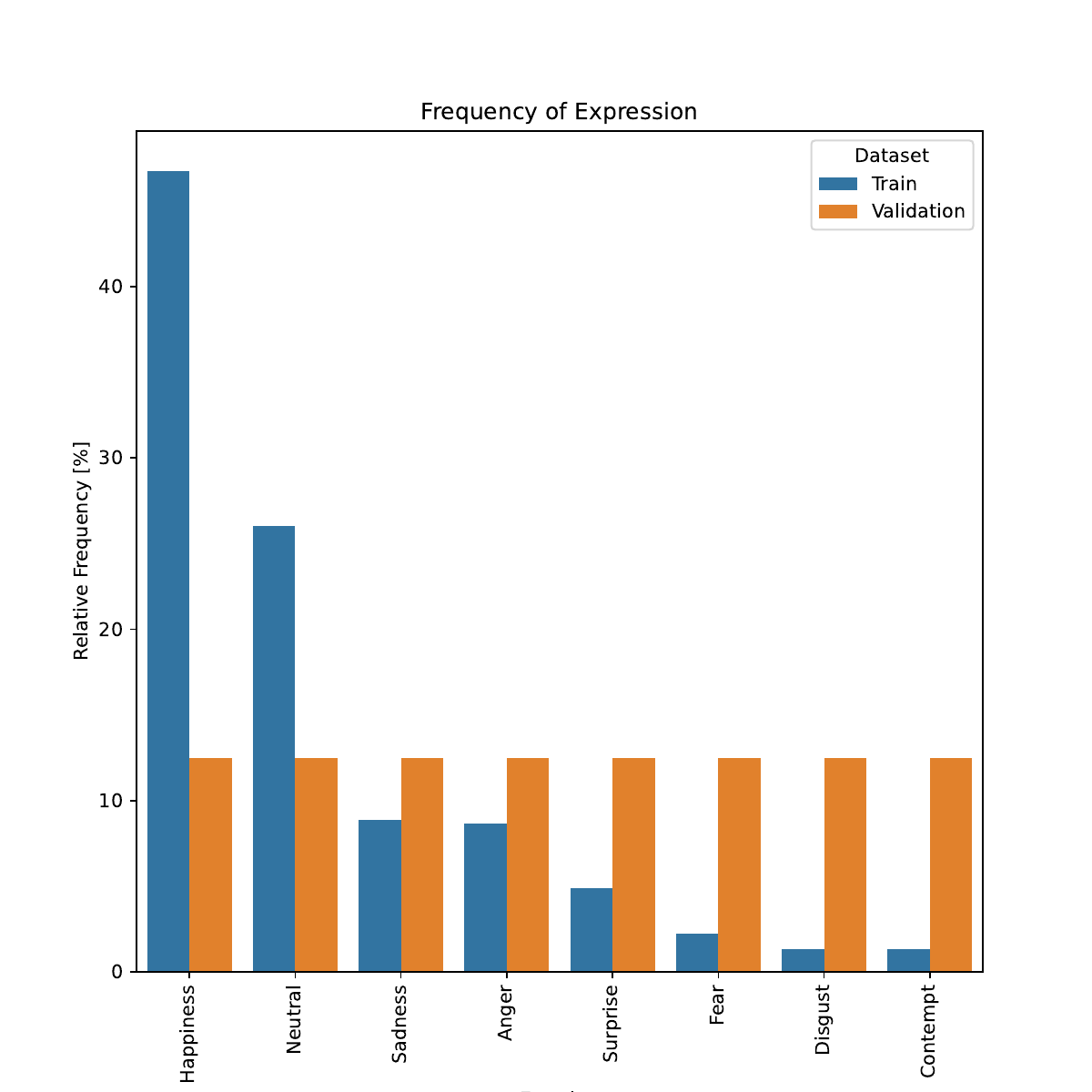}
    \caption{Frequency of expression of \affectnet{}.}
    \label{frequence\affectnet{}}
\end{figure}

\begin{figure}[t]
    \centering
    \includegraphics[width=\columnwidth, trim={4cm 0 2.5cm 0}, clip]{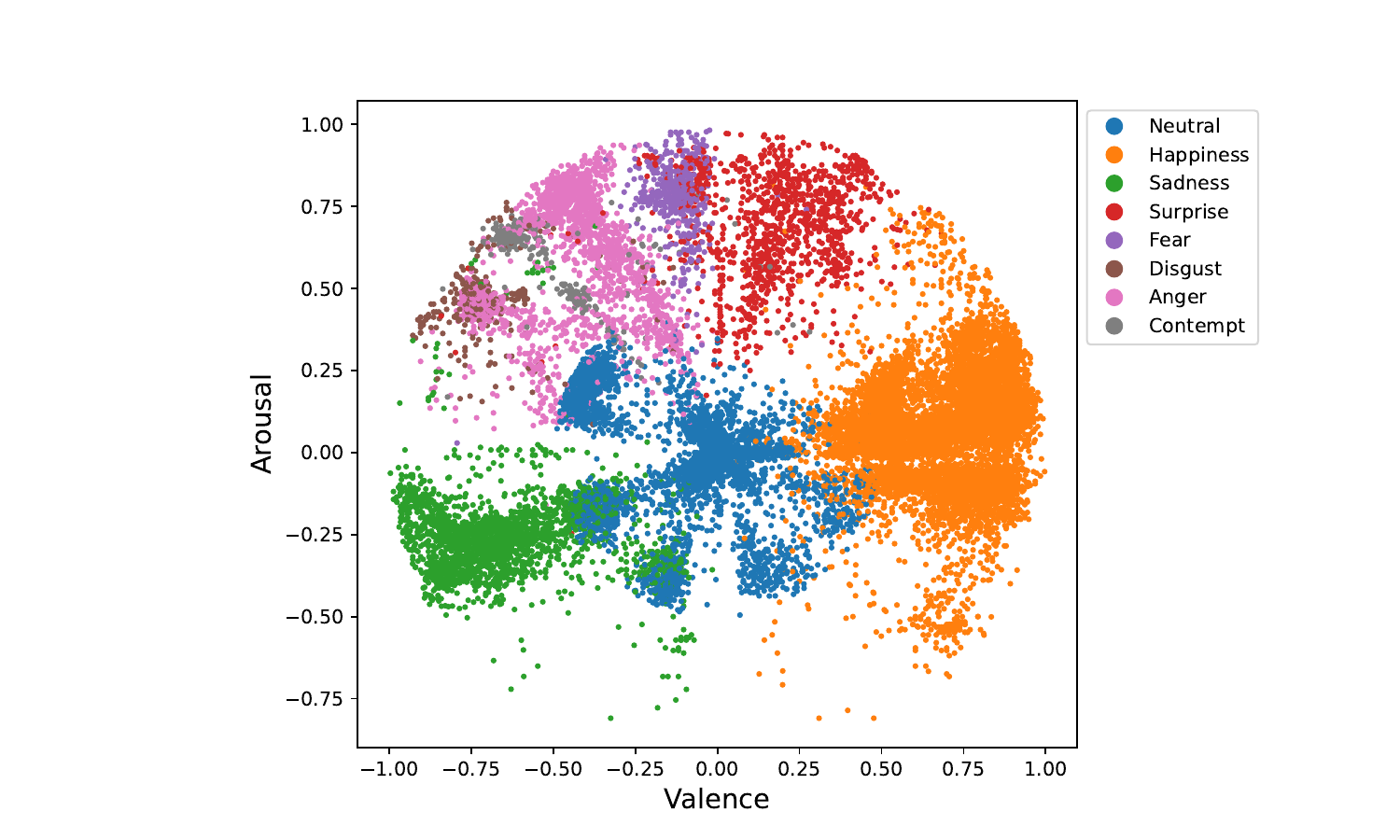}
    \caption{\VA{} in a subset of the train dataset of \affectnet{} sorted by expression category.}
    \label{fig:scatterplot}
\end{figure}

\subsection{\affectnet{}}

Images in the \affectnet{}~\cite{mollahosseini2017affectnet} dataset are labeled with (1) one out of eight possible discrete expression categories, (2) a \val{} value as a real number between -1 and 1, (3) an \aro{} value as a real number between -1 and 1, and (4) facial landmark points. The distribution of discrete categories (see Figure~\ref{frequence\affectnet{}}) is unbalanced. %, leading to a weighted loss function for model training. 
The sum of the probabilities would lead up to 70\% only with two of the eight expression categories (\textit{happiness} and \textit{neutral}). On the other hand, validation data is evenly distributed across all labels.

\begin{figure}[t]
    \centering
    \includegraphics[width=0.65\columnwidth]{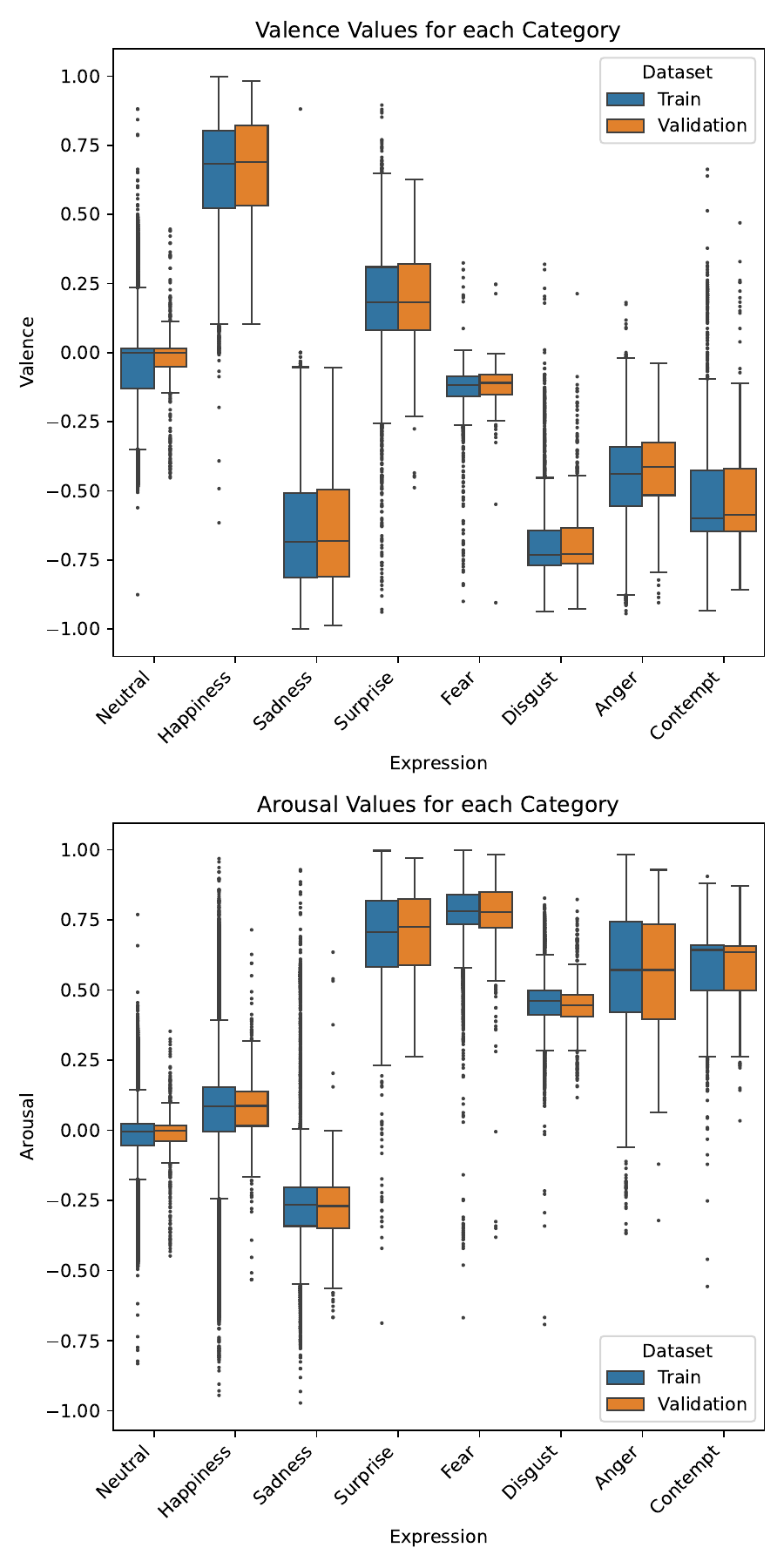}
    \caption{\VA{} in \affectnet{}-8.}
    \label{fig:affectnet_av_for_each_category}
\end{figure}

To analyze the distribution of the continuous values \va{}, we displayed the values from the training set in the circumplex model of affect as originally proposed by Russell~\cite{rusellmodell}, with the values of \aro{} on the ordinate and \val{} on the abscissa (see Figure~\ref{fig:scatterplot}). As a result, the visualization clearly reveals that different expression categories can lead to an overlap in the \val{}/\aro{} values. Additionally, we analyze the distribution of \val{}/\aro{} per category as shown in  Figure~\ref{fig:affectnet_av_for_each_category}. For example, \textit{neutral} and \textit{happiness} expressions share a similar median in \aro{} dimension, whilst having a different median in the \val{} dimension. As expected, the \textit{neutral} category is centered around zero for \va{}. 

% \begin{figure}[ht]
%     \centering
%     \includegraphics[width=\columnwidth]{pictures/affectnet/scatterplot.pdf}
%     \caption{Valence and Arousal sorted by Category}
%     \label{fig:scatterplot}
% \end{figure}

%\input{pictures/affectnet/frequency_of_expression.pgf}

A comparison of the \va{} values across the train and validation datasets shows that imbalance is also present.
\begin{figure}
    \centering
    \includegraphics[width=0.65\columnwidth]{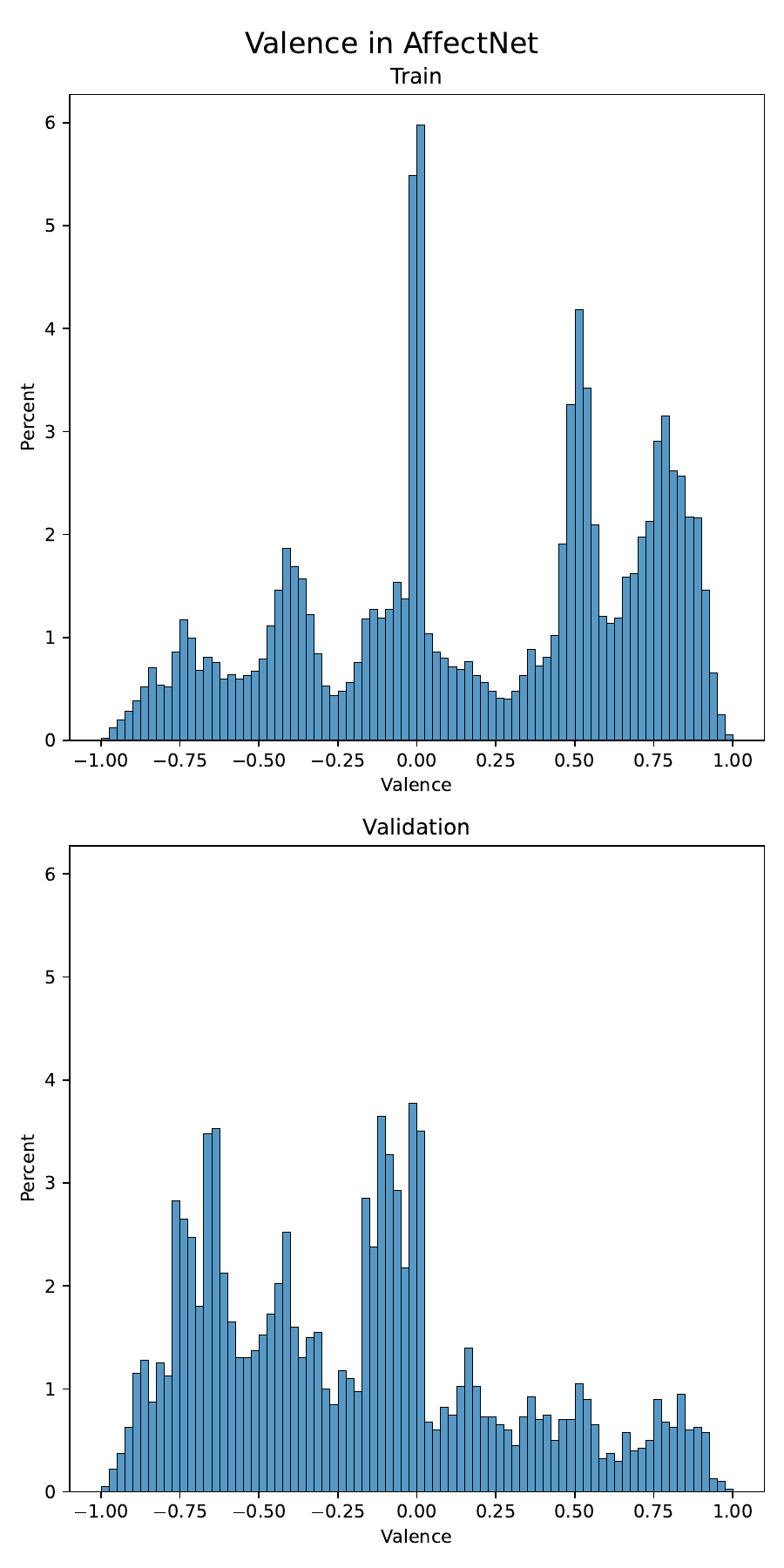}
    \includegraphics[width=0.65\columnwidth]{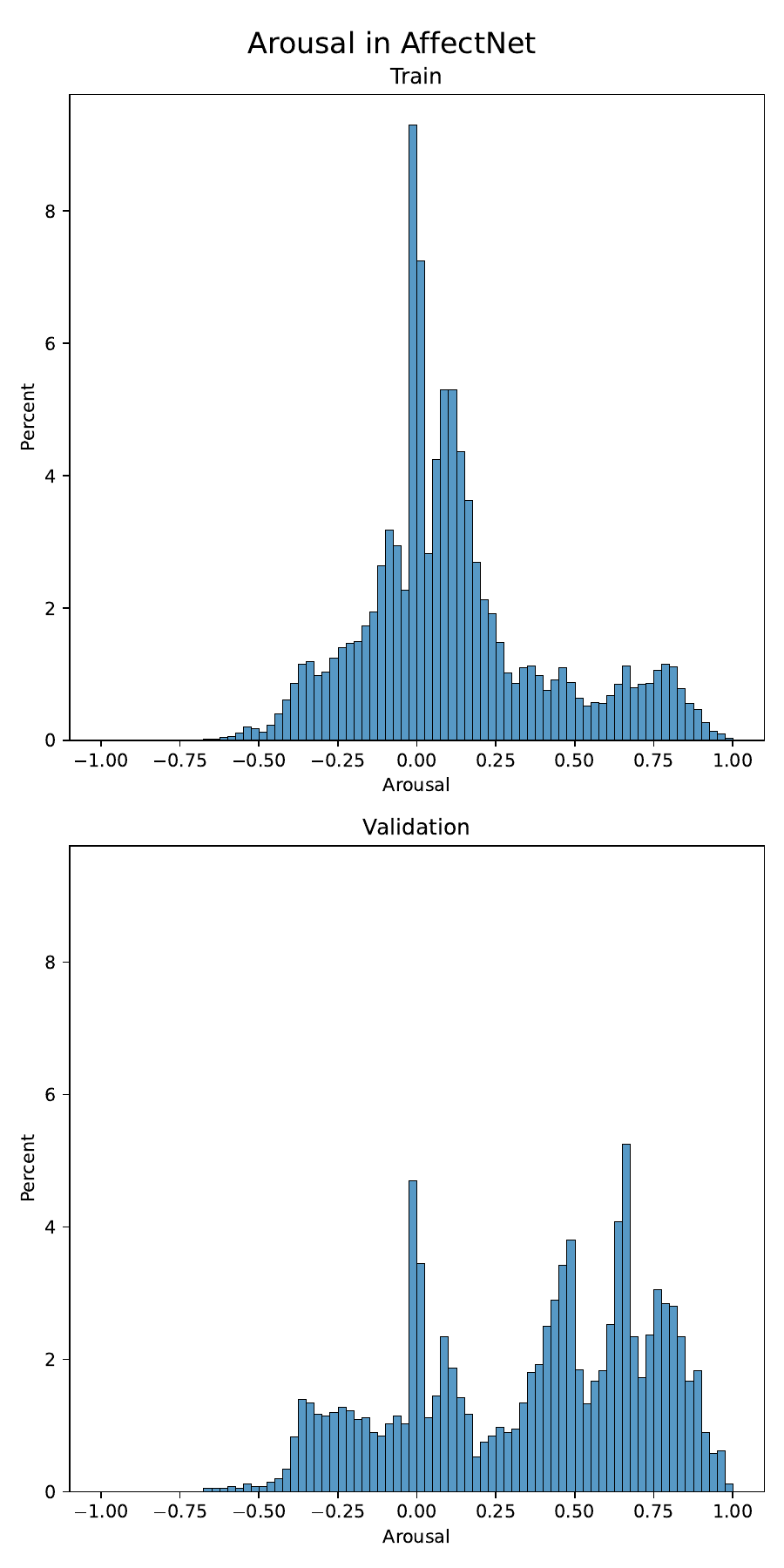}
    \caption{\Val{} and \Aro{} in \affectnet{}-8.}
    \label{fig:valence_distribution}
\end{figure}
In particular, more values from the training dataset are positive compared to the validation dataset with a higher portion of negative values (see Figure~\ref{fig:valence_distribution}). This imbalance is even more noticeable for arousal values. In the validation dataset, there are far more high-valued positive values compared to the training dataset.

\subsection{\emotic{}}
In \emotic{}~\cite{kosti_emotic_2017},  every image has a more complex labeling, targeting an overall context and a body focusing on the expression (see Figure~\ref{fig:example-imgs}). Each bounding box of a human is labeled with at least one expression, a \val{} value (integer between 1 and 10), an \aro{} value (integer between 1 and 10), a \dom{} value (integer between 1 and 10), gender (female/male) and age of a person (kid/teenager/adult).

% \begin{itemize}
%     \item Bounding box (Bbox) of the body image ([$x_1$ $y_1$ $x_2$ $y_2$])
%     \item Multi-labeling: At least one emotion
%     \item A \val{} value (integer between 1 and 10)
%     \item An \aro{} value (integer between 1 and 10)
%     \item A \dom{} value (integer between 1 and 10)
%     \item 
%     \item Age of the person in the Bbox (kid, teenager or adult)
% \end{itemize}

\newcolumntype{C}{>{\centering\arraybackslash}X} 

\begin{figure}[ht]
    \centering
    \includegraphics[width=\columnwidth]{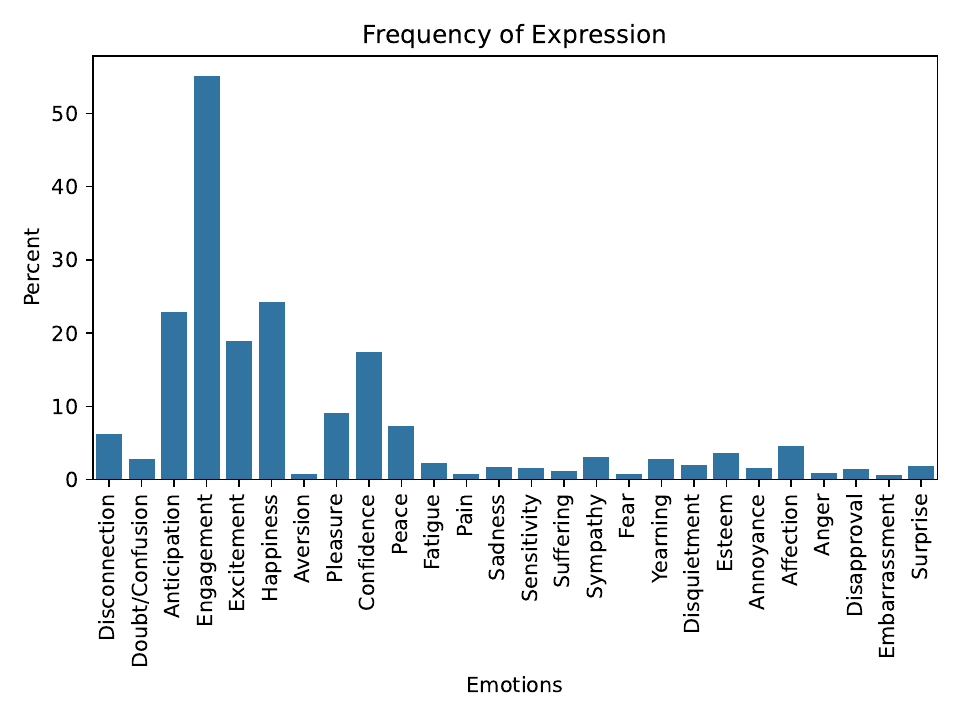}
    \caption{Distribution of expression frequency in \emotic{} train data.}
    \label{fig:emotic_labeldistr}
\end{figure}

Figure~\ref{fig:emotic_labeldistr} shows the relative occurrence of each expression in the dataset. Due to the multi-labeling of the dataset, an image can have multiple labels. Also, the overall frequency of the expression \textit{engagement} is dominating. %, leading to a weight in our loss function. 
Furthermore, all categories with a relative occurrence over 10\% are  "positive", corresponding to a positive \val{} value. 

\begin{figure}[t]
    \centering
    \includegraphics[width=0.66\columnwidth]{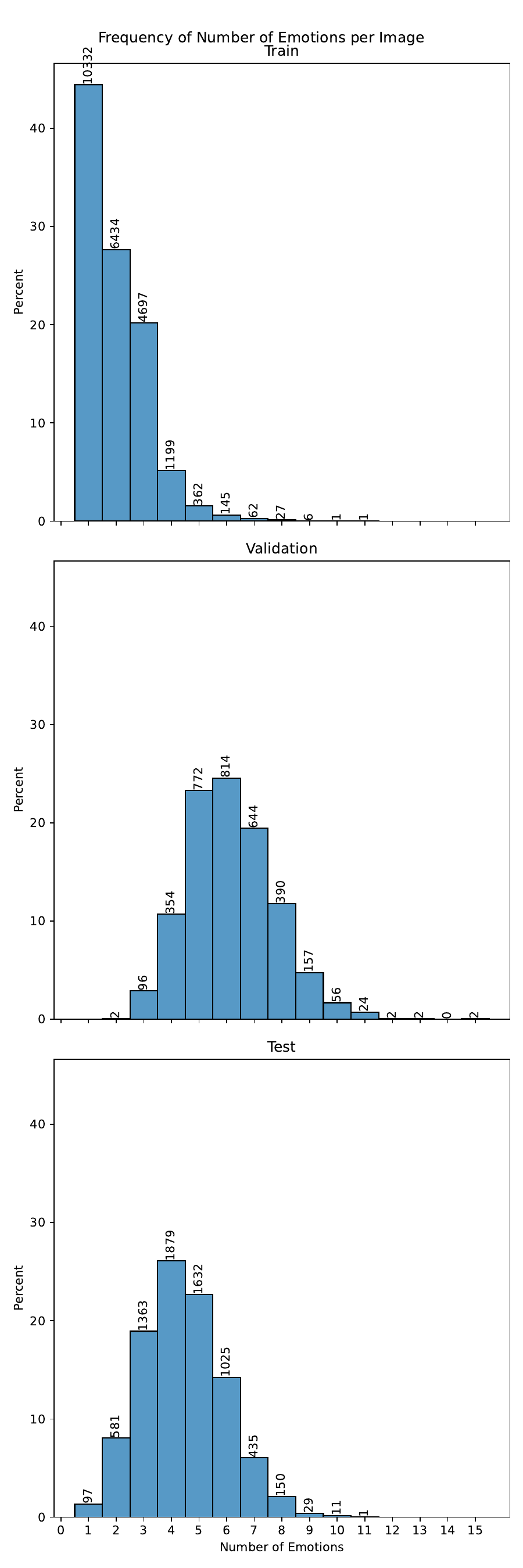}
    \caption{Instances of multiple true expressions per image in \emotic{} data.}
    \label{fig:emotic_label_occurance}
\end{figure}

Analysis of the label frequency in subsets has shown, that the training dataset contains a lot of images with one, two, or three categories, consistently decreasing. On the other side, the validation and the training dataset have a lot of images labeled with four or more categories  (see Figure~\ref{fig:emotic_label_occurance}). 

In summary, the \emotic{} dataset leads to a much more challenging task to train a suitable computer vision model. A fairly small dataset size, multi-person context, multi-label encoding, inconsistent unbalanced datasets, and different expression frequencies have led us to much more severe effort in the choice of suitable model hyperparameters.

% \newpage
% \clearpage
\section{Models for Discrete and Continuous Expression Inference}
\label{sec:method}

Starting with a state-of-the-art baseline model trained on the \affectnet{} data, we evaluated different approaches to check if combining the classification of discrete emotional expressions and regression for continuous \va{} values can lead to better results. After training and comparing our approaches on \affectnet{}, we use the architecture of the best model to train on the \emotic{} data.

\subsection{Baseline Selection and Losses}
 
%\subsection{Losses for \affectnet{}}
% Eigentlich wäre es besser wir hätten hier ein veröffentlichtes Paper nicht unser eigenes "custom-Student" B4 Netz 
To assess the impact of the usage of \val{}/\aro{} during training, we consider three model versions:

\textbf{Discrete models} with a cross-entropy loss. Due to an unbalanced class distribution, we used a weighted cross-entropy loss $L_{WeightedCE}$. The weights in the cross-entropy loss were set to the frequencies of expressions in the training dataset (see Table~\ref{tab:weights}). 

\textbf{Combined models} with an additional MSE loss for \va{}, weighted with a balance factor $\alpha$: 
\begin{equation}
L_{combined} = L_{WeightedCE} + \alpha \cdot L_{MSE}
\end{equation}

\textbf{Valence-arousal models} with a CCC loss for the regression of continuous valence and arousal values:

\begin{equation}
L_{valence-arousal} = L_{CCC} + \beta \cdot L_{MSE}
\end{equation}

\begin{table}[h]
\centering
\begin{tabular}{r | c | c}
\hline
 \textbf{Label} & \textbf{\affectnet{}-8} & \textbf{\affectnet{}-7} \\ 
  \hline
 Neutral & 0.015605 & 0.022600\\
 Happiness & 0.008709 & 0.012589\\
  Sadness & 0.046078 & 0.066464\\
 Surprise & 0.083078 & 0.120094\\
 Fear & 0.185434 & 0.265305\\
 Disgust & 0.305953 & 0.444943\\
 Anger & 0.046934 & 0.068006\\
 Contempt & 0.308210 & / \\ \hline
\end{tabular}
\caption{Weights for the cross-entropy loss.}
\label{tab:weights}
\end{table}

% \begin{equation}
% L_{CCC}(x, y) = -\frac{1}{N} \sum_{i=1}^{N} \frac{2 \cdot \text{covar matrix}}{\text{denominator}}
% \end{equation}

% Where:
% \[
% \text{covariance matrix } \Sigma = \frac{{(x - \mu_x)^T \cdot (y - \mu_y)}}{{N - 1}}
% \]
% \[
% \text{denominator} = \sigma^2_x + \sigma^2_y + (\mu_x - \mu_y)^2
% \]

%Our overall approaches can be categorized as follows:
% Hier müssen wir irgendwo die Modelarchitecture visualisieren

%/
%\begin{itemize}
%\item \textbf{Only Classification}: Starting from our transfer %learning approach, evaluated on the \affectnet{}-8 test datataset, %we utilized a new model architecture, changed the baseline data %augmentation, calculated a  weighted CrossEntropyLoss %($L_{WCE}$), optimized the hyperparemeters and use a  MaxViT-Tiny %architecture ( \textbf{Ändern !!!} )
%\item \textbf{Regression (Val \& Aro)}: With this approach, we %changed the model structure and our loss function to predict only %the two continuous values \va{} using a MSELoss.
%\item \textbf{Combined}: This approach combines the %classification- and regression approach. To incorporate this, the %model structure is extended to have 9/10 output neurons, splitted %in 7/8 classification neurons and two regression neurons. Hence %the overall loss for our combined approach is calculated using %the sum of the weighted CrossEntropyLoss ($L_{WeightedCE}$) and %the MSELoss ($L_{MSE}$), balanced by a factor $\alpha$. 
%\begin{center}
%\begin{math}
%L = L_{WeightedCE} + \alpha * L_{MSE}
%\end{math}
%\end{center}
%\end{itemize}

\subsection{Training Setup}
 
%With our defined model performance metrics, we present in Table~\ref{tab:metrics} our scores resulting from our different approaches. 
The models proposed above were trained on the AffectNet data. Then, the best-performing model was selected for retraining on EMOTIC data. All training was performed using NVIDIA 4090 GPUs. Table~\ref{tab:hyperparameters} shows the hyperparameters.

\begin{table}[htbp]
\centering
\begin{tabular}{r | l }
\hline
 \textbf{Hyperparameter} & \textbf{Value} \\  \hline
 Batch Size & 128 \\
 Learning rate & 5e-5 \\
  Optimizer & AdamW~\cite{loshchilov2017decoupled} \\
 Learning rate scheduler & Cosine annealing~\cite{loshchilov2016sgdr} \\
\textit{L}$_{combined}$  factor& $\alpha = 5$ \\
\textit{L}$_{valence-arousal}$ factor & $\beta = 3$ \\ \hline
\end{tabular}
\caption{Hyperparameters for model training.}
\label{tab:hyperparameters}
\end{table}

To train the proposed model architecture on \affectnet{}, we used the following data augmentation techniques:
\begin{itemize}
\item \textit{RandomHorizontalFlip} with p=0.5,
\item \textit{RandomGrayscale} with p=0.01,
\item \textit{RandomRotation} with degree=10, 
\item \textit{ColorJitter} with brightness=0.2, contrast=0.2, saturation=0.2 and hue=0.1,
\item \textit{RandomPerspective} with distortion=0.2 and p=0.5,
\item \textit{Normalize} with mean=[0.485, 0.456, 0.406] and std=[0.229, 0.224, 0.225],
\item \textit{RandomErasing} with p=0.5, scale=(0.02, 0.2), ratio=(0.3, 3.3) and value='random'.
\end{itemize}

Whilst most augmentation techniques target a more robust/stable model, we discovered that \textit{RandomErasing} prevented model overfitting on the training dataset, which would otherwise occur due to the small dataset size. Based on the training behavior, we have chosen a comparably high batch size and a relatively small learning rate. We noticed, that even with this small learning rate, the proposed model achieved the best results in the fifth epoch. However, a change in the model architecture (more/less parameters, change in model architecture, different batch size, etc.) did not improve our results.

\subsection{Evaluation Setup}

\textbf{Performance metrics for \affectnet{}:} to address the dual nature of the proposed model, which integrates a classification- and/or a regression task, we evaluate its performance using state-of-the-art binary classification metrics as well as common regression metrics: precision P, recall R, F1 score F1, mean absolute error (MAE), mean squared error (MSE), and root mean squared error (RMSE).

\textbf{Performance metrics for \emotic{}:} as mentioned above, we use the best model trained on \affectnet{} to retrain our model on the \emotic{} dataset. Because the \emotic{} dataset is a multi-label multi-classification dataset, a change of the loss is necessary. Hence, we changed the cross-entropy loss to a positive-weighted binary cross-entropy (BCE) loss, where the positive weights are defined as the inverse of the occurrence of each label.
%ratio between a label is present versus not for each class.

\begin{center}
\begin{math}
\tilde{L}_{combined} = L_{WeightedBCE} + \alpha \cdot L_{MSE}
\end{math}
\end{center}

\textbf{Cross-Validation of models:} as both \emotic{} and \affectnet{} share the same dimension regarding \va{}, we test the proposed trained models on each test data. To achieve this, we transformed the dimension of \va{} to ensure its values are between -1 and 1, then evaluated the datasets/models' generalization on thoroughly unseen data samples.

\begin{table*}[th]
  %\resizebox{1.0\linewidth}{!}{%
    \begin{tabular}{r  | l | c | c | c | c | c | c |c } %{\textwidth}{@{} l *{6}{C} c @{}}
    \hline
    \textbf{Dataset}& \textbf{Model}
    & \textbf{Precision $\uparrow$} & \textbf{Recall $\uparrow$} & \textbf{F1 $\uparrow$} & \textbf{MSE $\downarrow$} & \textbf{MAE $\downarrow$} & \textbf{RMSE $\downarrow$} & \textbf{CCC $\uparrow$} \\\hline

     & EfficientNetv2s$_{discrete}$  & 0.634 & 0.631 & 0.631 & - & - & - &- \\
     & MaxViT$_{discrete}$  & 0.640 & 0.639 & 0.638 & - & - & -&- \\\cline{2-9}
       \affectnet{}-7   & EfficientNetv2s$_{combined}$ & 0.650 & 0.646 & 0.646 & 0.0956 & 0.2298 & 0.3092  & 0.7636\\
  & MaxViT$_{combined}$ & \textbf{0.666} & \textbf{0.664} & \textbf{0.664} & 0.0947 & 0.2251 & 0.3077  & 0.7640\\\cline{2-9}
    & EfficientNetv2s$_{valence-arousal}$ & - & - & - & \textbf{0.0833} & \textbf{0.2098} & \textbf{0.2887} & \textbf{0.8206} \\
    & MaxViT$_{valence-arousal}$ & - & - & - & 0.0841 & 0.2121 & 0.2901 & 0.8196 \\ \hline

    %%%%%%%AffectNet-8
     & EfficientNetv2s$_{discrete}$ & 0.605 & 0.599 & 0.599 & - & - & - &-\\
     & MaxViT$_{discrete}$ & 0.602 & 0.598 & 0.599 & - & - & - &-\\ \cline{2-9}
    \affectnet{}-8 & EfficientNetv2s$_{combined}$ & 0.612 & 0.606 & 0.607 & 0.1420 & 0.2781 & 0.3769 & 0.6413\\  
    & MaxViT$_{combined}$ & \textbf{0.623} & \textbf{0.621} & \textbf{0.621} & 0.1370 & 0.2715 & 0.3701 & 0.6592\\\cline{2-9}
      & EfficientNetv2s$_{valence-arousal}$ & - & - & - & 0.1028 & 0.2387 & 0.3206 & 0.7816\\ 
    & MaxViT$_{valence-arousal}$ & - & - & - & \textbf{0.1021} & \textbf{0.2351} & \textbf{0.3196} & \textbf{0.7840} \\
    %\addlinespace
    
    \hline
    \end{tabular}
%}
\caption{Comparison of model performance on \affectnet{}. The best results for \affectnet{}-7 and \affectnet{}-8 are highlighted.}
\label{tab:metrics}
\end{table*}

\section{Evaluation}
In the following, we discuss and compare model performance on \affectnet{} and \emotic{}. 

\subsection{Model Architecture}

For comparison, we have evaluated both CNN- and transformer-based architectures, while focusing on lightweight models: EfficientNetv2~\cite{tan2021efficientnetv2} and %MobileNetv2~\cite{sandler2018mobilenetv2}, 
MaxViT-Tiny ~\cite{tu2022maxvit}~\cite{pytorchmaxvit}. Furthermore, we have experimented with Swin transformer~\cite{liu2021swin} models. However, these have demonstrated worse results with precision below $0.35$. PyTorch implementations of models, pre-trained on ImageNet~\cite{deng2009imagenet} were used.  The best results were achieved with the MaxViT models (see Table~\ref{tab:metrics}).

\subsection{Impact of Training with Valence and Arousal on Discrete Expressions for \affectnet{}}

A different model architecture and a combined training approach increased the baseline F1-score from 60\% to 62\% when using the \affectnet{}-8 dataset (see Table~\ref{tab:metrics}). With a reduced \affectnet{}-7 dataset, we also increased our model performance leading to an F1 score of 66\%.  The combined approach thus improved the classification results for both datasets by 2\%. 

\begin{figure}[b]
    \centering
    \includegraphics[width = 0.9\columnwidth]{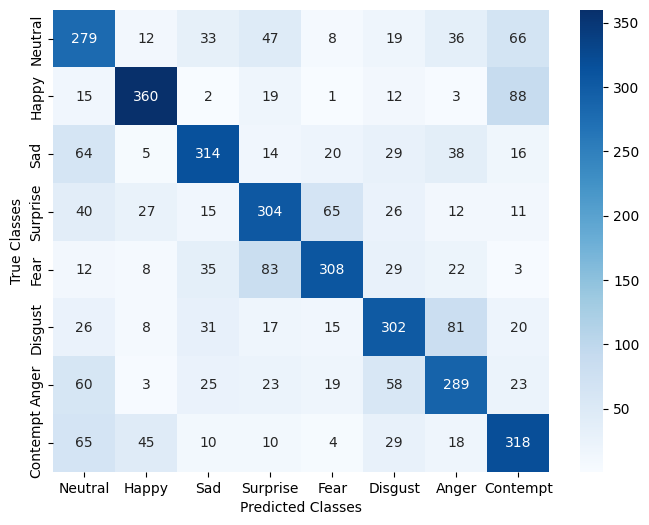}
    \caption{Confusion matrix for MaxViT$_{combined}$ on AffectNet-8.}
    \label{fig:confusion_matrix_classifier}
\end{figure}

Figure \ref{fig:confusion_matrix_classifier} displays the confusion matrix of the MaxViT$_{combined}$ for \affectnet{}-8. In accordance with the theory of the circumplex model of affect, the displayed transition of discrete emotional expressions is smooth. For example, \textit{surprise} and \textit{fear} have a similar median \aro{} value or \textit{disgust} and \textit{anger} around their corresponding \val{} value. A model using continuous values can thus potentially outperform the one with discrete values. %This led us to the evaluation of the regression performance. 

\subsection{Best \affectnet{} Model Regarding Valence and Arousal}
  
In contrast to the proposed combined training methodology, the best regression results were gained with the MaxViT$_{valence-arousal}$ model. To reduce noticeable oscillating behavior during training, we reduced the training dataset by balancing according to the discrete expression labels. Furthermore, we added the CCC loss to the L$_{valence-arousal}$ loss function and used the pre-trained weights of the best model from the combined method (MaxViT$_{combined}$). With a duration of two minutes per epoch, the best results were achieved in epoch seven.

\begin{table}[htbp]
\centering
\begin{tabular}{r | c | c | c}
\hline
\textbf{Metric} & \textbf{VGG-F}~\cite{bulat2022pretraining} & \textbf{Ours} & \textbf{Improvement} \\ 
\hline
RMSE$_{valence}$ $\downarrow$& 0.356 & \textbf{0.331} & 7,0\%\\
RMSE$_{arousal}$ $\downarrow$& 0.326 & \textbf{0.305} & 6,4\%\\
CCC$_{valence}$ $\uparrow$& 0.710 & \textbf{0.716} & 0,8\%\\
CCC$_{arousal}$ $\uparrow$& 0.629 & \textbf{0.642} & 2,0\%\\
\hline
\end{tabular}
\caption{Benchmark  vs. MaxViT$_{combined}$ for AffectNet-8}
\label{tab:benchmarkourapproach}
\end{table}

\begin{figure}[t]
    \centering
    \includegraphics[width=0.95\columnwidth]{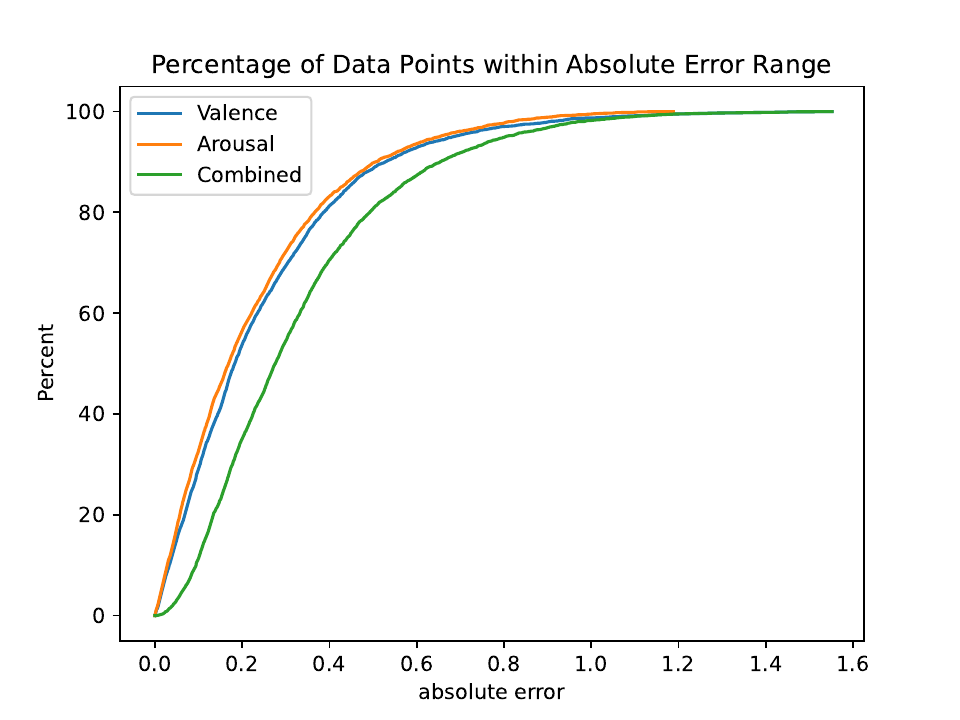}
    \caption{Absolute error for MaxViT$_{combined}$ trained on \affectnet{}-8.}
    \label{fig:affectnet_cdf}
\end{figure}

Figure~\ref{fig:affectnet_cdf} shows the percentage of data points within the absolute error range. When focusing on the ordinate, 80\% of the \va{} predictions differ only $\pm  0.3$ from their true value. The resulting model is thus more robust. %As a result, we are convinced that the learning approach led our training approach to a far robust model.

%\vspace{0.2cm}
\subsection{Performance of the \emotic{} Model}
 
For the \emotic{} dataset, we calculated the positive weights for each label for the training and validation dataset. Similar to the weights of the cross-entropy loss, the positive weight for each class is the inverse of the overall occurrence of a label. We chose this method to compensate for the imbalance of the frequency of expression (see Figure~\ref{fig:emotic_labeldistr}). 

 Table ~\ref{tab:benchmarkourapproachemotic} shows the overall metrics of our best EMOTIC model. To compare the RMSE of \va{} with the AffectNet dataset, we added a scaled version. For this, integers between 1 and 10 from EMOTIC are scaled to real values between -1 and 1. By definition, the CCC is invariant to shifts and scale transformations. Additionally, our EMOTIC model outperforms the best model by Khan \etal ~\cite{khan2024focusclip} according to Top-3 accuracy, which reaches 13.73 \% on the benchmark ~\cite{paperswithcodeemotop3}. 

\begin{table}[htbp]
\centering
\begin{tabular}{r | c | c}
\hline
\textbf{Metric} & \textbf{Ours} (Original) & \textbf{Ours} (Scaled) \\
\hline
Top-3 Accuracy & 14.73 \% & / \\ 
RMSE$_{valence}$ $\downarrow$& 1.150 & \textbf{0.256} \\ 
RMSE$_{arousal}$ $\downarrow$& 1.209 & \textbf{0.269}\\ 
RMSE$_{dominance}$ $\downarrow$& 1.169 & \textbf{0.260} \\ 
CCC$_{valence}$ $\uparrow$& 0.316 & \textbf{0.316} \\ 
CCC$_{arousal}$ $\uparrow$& 0.594 & \textbf{0.595} \\
CCC$_{dominance}$ $\uparrow$& 0.300 & \textbf{0.301}\\ \hline
\end{tabular}
\caption{Performance of MaxViT$_{combined}$ on EMOTIC.}
\label{tab:benchmarkourapproachemotic}
\end{table}

%Similar to our \affectnet{} model, we assessed the CDF performance of our model trained on the \emotic{} data. As shown in Figure~\ref{fig:emotic_cdf} our \emotic{} model generates similar results when assessing the body images of the test dataset. 
\subsection{Cross-Validation of the Models for Valence }

We assess the performance of the proposed trained \affectnet{}/\emotic{} model on the respective test datasets. The analysis of cross-validation results revealed that \affectnet{} outperformed \emotic{} notably in terms of absolute error metrics when evaluated on the \affectnet{} dataset. 

%\begin{figure}[t]
   %\centering
    %\includegraphics[width = \columnwidth]{pictures/inference_affectnet8_on_emotic.pdf}
    %\caption{CDF Absolute Error \affectnet{}-8 Model On \emotic{}}
    %\label{fig:affectnet8onemotic}
%\end{figure}

%\begin{figure}[t]
 %   \centering
  %  \includegraphics[width = \columnwidth]{pictures/inference_emotic_on_affectnet8.pdf}
   % \caption{CDF Absolute Error \emotic{} Model On \affectnet{}-8}
    %\label{fig:emoticonaffectnet8}
%\end{figure}

%\begin{table}[htbp]
%\centering
%\begin{tabular}{l | c | c }
%\textbf{Metric} & \textbf{\affectnet{} model} & \textbf{\emotic{} model} \\ 
%& \textbf{on \emotic{}} & \textbf{on \affectnet{}} \\ 
%\hline
%RMSE \val{} & 0.379 & 0.465 \\ 
%\hline
%RMSE \aro{} & 0.369 & 0.582 \\ 
%\hline
%CCC \val{} & 0.070 & 0.226 \\ 
%\hline
%CCC \aro{} & 0.056 & 0.007 \\ 
%\end{tabular}
%\caption{Cross Validation of our Best Models for \VA{} Estimation}
%\label{tab:benchmarkourapproachcrossvali}
%\end{table}

\begin{figure}[htbp]
    \centering
    \includegraphics[width=0.95\columnwidth]{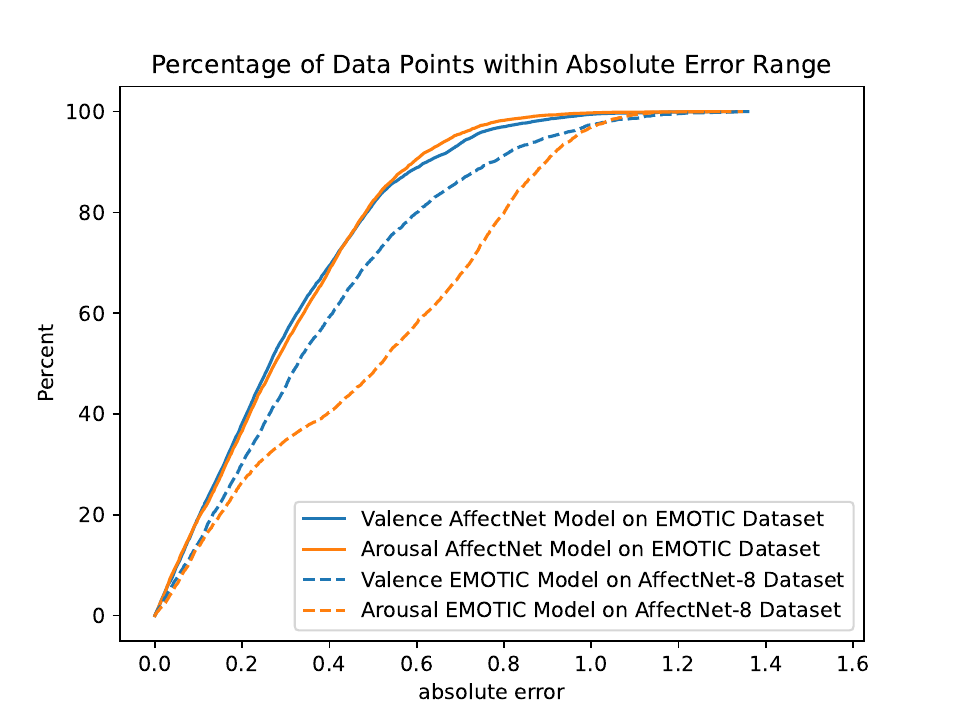}
    \caption{Absolute error for cross-validation of the MaxViT models.}
    \label{fig:benchmarkourapproachcrossvali}
\end{figure}

As Figure \ref{fig:benchmarkourapproachcrossvali} shows, the absolute errors for \va{} are significantly higher. Thus, the substantial advantage of \affectnet{} over \emotic{} in absolute error rates during cross-validation stresses its generalization ability for expression inference tasks.

\section{Conclusion \& Outlook}
\label{sec:conclusion}

In this paper, we assessed the capability of discrete classifier approaches with multi-task learning models when inferring emotional expressions. 
We used two prominent datasets tailored for discrete expressions and values based on the circumplex model of affect to train our models. 

\textbf{Firstly}, we have performed in-depth analysis of the datasets. It was observed that while test datasets are often balanced concerning emotional expressions, the balance is not maintained for \va{}. Models trained solely on \va{} tend to minimize errors. Additionally, it is noteworthy to delve into the intricate distribution of the \emotic{} dataset, especially how it varies concerning the number of classes in the train and test sets. 

\textbf{Secondly}, we proposed to use the MaxViT model architecture and described the training and evaluation protocol for both datasets. The proposed approach significantly improved model accuracy. Even in cases of misclassification, the predicted \va{} values often remained accurate. Establishing a threshold for correct prediction of \va{} poses an interesting challenge for future work, as it involves considering factors such as human error and the inherent complexity of emotional expression perception. Furthermore, our model based on \affectnet{} demonstrated robust performance in \va{} estimation via cross-validation. This suggests the potential for it to serve as a well-generalized model. Conversely, the performance of our \emotic{}-based approach was less conclusive, possibly due to insufficient data or other factors. 

In conclusion, our research underscores the effectiveness of continuous value approaches within multi-task learning frameworks for emotional expression inference. Further exploration and refinement of these methodologies could yield even more accurate and robust models in the future.

\newpage
\clearpage
{
    \small
    \bibliographystyle{ieeenat_fullname}
    \bibliography{main}
}


\end{document}